%% file: root.tex
\documentclass[conference]{ieeeconf}
\IEEEoverridecommandlockouts  
\usepackage{graphicx}
\usepackage[caption=false]{subfig}
\usepackage{amsmath}
\usepackage{amssymb}
\usepackage[acronym]{glossaries}
\usepackage{tikz}
\usepackage{tikz-3dplot}
\usepackage{array}
\usepackage{multirow}
\usepackage{hyperref}
\usepackage{cleveref}
\usepackage{url}
\usepackage{xcolor}
\usepackage{mathtools}
\usepackage{booktabs}
\usepackage[caption=false]{subfig}
\usepackage{svg}
\usepackage{footmisc}
\usepackage{fontawesome}

\usetikzlibrary{calc}
\usetikzlibrary{shapes.geometric}
\usetikzlibrary{fit}
\usetikzlibrary{positioning}
\usetikzlibrary{decorations.pathmorphing, decorations.pathreplacing,angles,quotes}
\usetikzlibrary{arrows}
\usetikzlibrary{shadows}
\usetikzlibrary{backgrounds}
\usetikzlibrary{calc,patterns}

\input{symbols.sty}
\input{definitions}

\newcommand\rurl[1]{%
  \href{https://#1}{\nolinkurl{#1}}%
}

\title{\LARGE \bf
A Framework for Real-World Multi-Robot Systems\\Running Decentralized GNN-Based Policies
}

\author{Jan Blumenkamp, Steven Morad, Jennifer Gielis, Qingbiao Li and Amanda Prorok
\thanks{All authors are with Department of Computer Science and Technology, University of Cambridge.
       Email: {\tt\footnotesize \{jb2270, sm2558, jag233, ql295, asp45\}@cam.ac.uk}.}%
}

\begin{document}

\maketitle
\thispagestyle{empty}
\pagestyle{empty}

\setlength{\textfloatsep}{10pt}


\begin{abstract}
\Glspl{gnn} are a paradigm-shifting neural architecture to facilitate the learning of complex multi-agent behaviors. Recent work has demonstrated remarkable performance in tasks such as flocking, multi-agent path planning and cooperative coverage. However, the policies derived through \Gls{gnn}-based learning schemes have not yet been deployed to the \gls{real_world} on physical multi-robot systems. In this work, we present the design of a system that allows for fully decentralized execution of \Gls{gnn}-based policies. We create a framework based on ROS2 and
elaborate its details in this paper. We demonstrate our framework on a case-study that requires tight coordination between robots, and present first-of-a-kind results that show successful \gls{real_world} deployment of GNN-based policies on a decentralized multi-robot system relying on \gls{adhoc} communication. A video demonstration of this case-study can be found online\footnote{\label{footnote:video}\rurl{youtube.com/watch?v=COh-WLn4iO4}}.
\end{abstract}

\glsresetall 

\section{Introduction}
Significant effort has been invested into finding analytical solutions to multi-robot problems, balancing optimality, completeness, and computational efficiency~\cite{barer2014suboptimal, van2005prioritized, standley2011complete,wu_MultiRobot_2019}.
Data-driven approaches can find near-optimal solutions to NP-hard problems, enabling fast on-line planning and coordination, as typically required in robotics. This has thus provided alternatives for the aforementioned challenges ~\cite{sartoretti_PRIMAL_2019,khan_2020,li2021message,zhou2021graph}.
\Glspl{gnn}, in particular, demonstrate remarkable performance and generalize well to large-scale robotic teams for various tasks such as flocking, navigation, and control~\cite{tolstaya_2020,khan_2020,li_2020, kortvelesy2021modgnn, morad2021graph, blumenkamp_2020}. In such multi-robot systems, \glspl{gnn} learn inter-robot communication strategies using latent messages. Individual robots aggregate these messages from their neighbors to overcome inherently local (partial) knowledge and build a more complete understanding of the world they are operating in.

While \Gls{gnn}-based policies are typically trained in a centralized manner in \gls{sim}, and therefore assume synchronous communication, resulting policies can be executed either in a centralized or decentralized mode. Evaluating a \Gls{gnn} in the \emph{centralized} mode typically requires execution on a single machine decoupled from the robots that are acting according to the policy \cite{khan_2020, li_2020, Jan_2021_ICRA2021}. This \emph{(i)} introduces a single point of failure, \emph{(ii)} requires all robots to maintain constant network connectivity, and \emph{(iii)} introduces scalability issues due to computational complexity $\ccalO(N^2)$ where $N$ is the number of robots. In contrast, in the \emph{decentralized} mode, each robot is responsible for making its own decisions. With fully decentralized evaluation, \emph{(i)} there is no single point of failure, resulting in a higher fault tolerance, \emph{(ii)} agents do not need to remain in network range of a router that orchestrates the evaluation, and \emph{(iii)} computation is parallelized across $N$ robots, decoupling the time complexity from the number of robots. 

\begin{figure}[tb]
    \centering
    \includegraphics[width=\columnwidth]{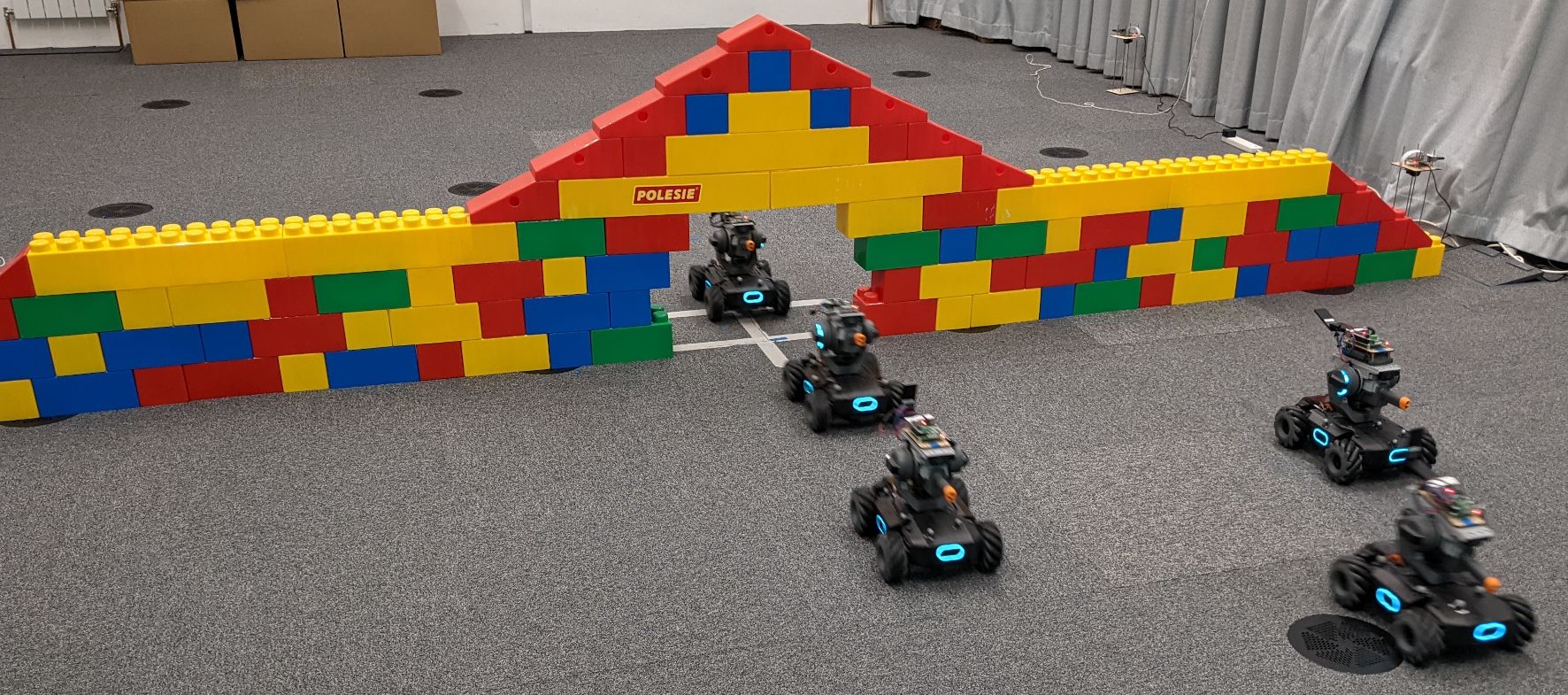}
    \caption{We deploy a set of five DJI RoboMaster robots in a real-world setup using \Glspl{gnn} and \gls{adhoc} communication. The robots navigate through a narrow passageway to reconfigure on the other side, as quickly as possible. \label{fig:hero_figure}}
\end{figure}

Even though \Glspl{gnn} have an inherently decentralizable mathematical formulation, previous work on \Gls{gnn}-based multi-robot policies was conducted exclusively in centralized simulations using synchronous communication~\cite{li_2020, kortvelesy2021modgnn, tolstaya_2020}. For practical reasons, decentralized execution is often unavoidable in the \gls{real_world}, but it is currently unknown whether this contributes to a shift of domains, and how resulting policies are affected.
Multi-robot \Glspl{gnn} require inter-robot communication, but \gls{real_world} wireless communication is noisy, and messages can be lost or delayed, leading to significant performance loss---this is exemplified in prior work that demonstrates the need for appropriate models to overcome these challenges
\cite{calvo2021ros, gregory2016enabling,stephan2017concurrent, stephan2014robust}.
Further compounding these issues, decentralized policies are typically executed asynchronously, resulting in system states not previously encountered during training. 

In this paper, we provide a framework that facilitates the decentralized execution of \Gls{gnn}-based multi-robot policies. We present the results of a suite of real-robot experiments (see \autoref{fig:hero_figure}) to demonstrate the consequences of this decentralized execution. To that end, we introduce a taxonomy of different evaluation modes and networking configurations.
Specifically, we contribute:
\begin{enumerate}
    \item A ROS2-based software and networking framework for \Glspl{gnn} and other message-passing algorithms to facilitate operation in both \gls{sim} and the \gls{real_world}, and to permit GNN-based policy execution in either a centralized or decentralized manner. We provide the source code online.\footnote{\rurl{github.com/proroklab/ros2_multi_agent_passage}}
    \item An ablation study on several forms of execution to quantify performance shifts between centralized execution and three forms of decentralized policy execution, \emph{(i)} offboard (non-local), \emph{(ii)} onboard over routing infrastructure, and \emph{(iii)} onboard with \gls{adhoc} networking.
\end{enumerate}

\section{Related Work}
In this section, we first review related multi-robot systems testbeds and frameworks. Our survey includes centralized frameworks as well as decentralized methods that either use machine-learning based approaches or communication. We emphasize that none of these methods combine learning-based methods and communication. Lastly, we review the related work on robotic communication frameworks and standards to evaluate an appropriate choice for our use-case.

\textbf{Multi-Robot Systems Testbeds.}
Remotely accessible mobile and wireless sensor testbeds are in high demand both in research and industry. Mobile Emulab~\cite{johnson2006mobile} and CrazySwarm~\cite{James2017} were developed as centrally controlled \gls{real_world} multi-robot research platforms. 
As decentralized platforms gained popularity, roboticists developed a variety of systems for small footprint robot swarms, including Robotarium~\cite{pickem2017robotarium}, Micro-UAV~\cite{michael2010grasp} or IRIS~\cite{tran2018intelligent} to large scale platforms such as HoTDeC~\cite{stubbs2006multivehicle}. 
These platforms provide testbeds for decentralized control and communication. However, none of these systems utilize machine-learning-based policies, and only few learning-based methods have demonstrated \gls{real_world} experiments~\cite{sartoretti_PRIMAL_2019}. 
Although work at the intersection of machine-learning and multi-robot control shows remarkable performance~\cite{sartoretti_PRIMAL_2019, khan_2020,li_2020, prorok2021holy, wang2020mobile, zhou2020smarts}, little work has been done to show how to make these methods practical (i.e., \gls{real_world}).
Of particular interest is how explicit inter-robot communication~\cite{tolstaya_2020, li_2020, foerster_2016} plays a role in accumulating information from other robots.
A recent study investigates the robustness of decentralized inference of binary classifier GNNs in wireless communication systems~\cite{lee2021decentralized}, but their work is limited to simulation and does not focus on communication contention and latency.
These learning-based multi-agent platforms and multi-robot frameworks are either restricted to \gls{sim}~\cite{khan_2020, li_2020,wang2020mobile}, rely on centralized evaluation~\cite{sartoretti_PRIMAL_2019,Jan_2021_ICRA2021}, or are only evaluated in simulated experiments for decentralized wireless communication. There is a gap between simulation-based testbeds and testbeds that facilitate the deployment of policies derived from machine-learning methods to the \gls{real_world}.

\textbf{Robotic Communications Frameworks.}
Communications between agents and controllers is a ubiquitous requirement on experimental robotics platforms, either for experimental control or operational messaging. For these functions, the IEEE 802.11 (commonly WiFi) and 802.15 protocol suites are commonly used \cite{frotzscher2014}, with various communications frameworks are overlaid on top of these low-level technologies (e.g. RTPS, MQTT \cite{tran2018intelligent} or standard IP \cite{stubbs2006multivehicle}).
Whatever the specific technology, the underlying protocol suites and the nature of wireless communication set fundamental limitations \cite{wang2006_80211} on available messaging rates when multiple agents are communicating in a decentralized manner. Multiple strategies exist that attempt to maximize protocol performance under specific conditions \cite{bhargava2018collisions, gielis2021collisions}, including dynamic centralization using homogeneous agents \cite{kang2019recen}. Despite these strategies, the performance of these systems at scale remain poorly tested in \gls{real_world} robotics systems, which often entail unexpected overheads \cite{kronauer2021latency}.

\section{Preliminaries}
In this section, we review the formalization of \Glspl{gnn} as well as the basic functionalities of \Gls{ros}, the software library that we build on.

\begin{figure}[tb]
    \centering
    \resizebox{\columnwidth}{!}{\input{tikz_figs/hero.tex}}
    \caption{The robots form a graph based on their separation and communication range $R_{COM}$. They leverage communication via latent messages $\bbm^{i,t}$ generated from local observations $\bbz^{i,t}$ propagated over graph edges (wireless \Gls{adhoc} communication links) to overcome the partial observability of the workspace. To solve this task, we utilize and deploy \Gls{gnn}-based policies that aggregate messages of robots within the local neighborhood $\ccalN^{t,i}$ and compute a local action.  \label{fig:gnns}}
\end{figure}
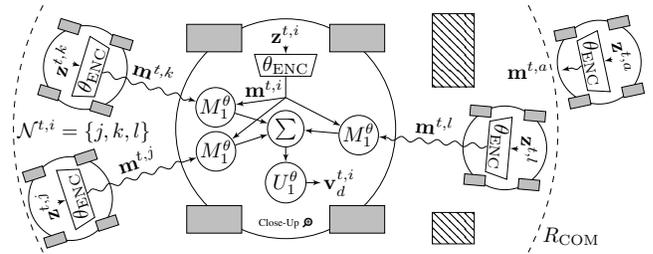

\subsection{Graph Neural Networks}
\label{sec:gnn}
A multi-robot system can be defined as graph $\ccalG = \langle \ccalV, \ccalE \rangle$, where each robot is represented as a node in the node set $\ccalV = \{1, \dots, n\}$. The inter-robot relationships are represented as edge set $\ccalE = \ccalV \times \ccalV$ with edge features $\bbe^{t,ji} \in \ccalE$ at each discrete time step $t$. If robot $j$ is in communication range $R_\mathrm{COM}$ of robot $i$, it is in robot $i$'s neighborhood $j \in \ccalN^{t, i}$ and robot $i$ can emit a message $\bbm^{t,i}$ that is broadcast to its neighbors.

Neural message passing \cite{gilmer2017neural} updates the hidden state $\bbh_k^{t+1,i}$ of each robot $i$ for each neural network layer $k$ using the message function $M$ and the vertex update function $U$ according to
\begin{equation}
    \small
	\bbh_k^{t+1,i} = U_k^\theta\left(\bbh_{k-1}^{t,i}, \textstyle \sum_{j\in \ccalN^{t,i}} M_k^\theta\left(\bbh_{k-1}^{t,i},\bbh_{k-1}^{t,j},\bbe^{t,ji}\right)\right),
	\label{eq:message_passing}
\end{equation}
where $U$ and $M$ are functions with learnable parameters $\theta$. The decentralized evaluation is explained in \autoref{fig:gnns}. Although centralized formulations also exist, according to \eqref{eq:message_passing}, evaluating a \Gls{gnn} is a fully decentralizable operation depending only on received messages and local information.

\subsection{ROS and ROS2}
\Gls{ros} is a set of open-source libraries for messaging, device abstraction, and hardware control \cite{quigley2009ros}. ROS generates a peer-to-peer graph of processes (\emph{Nodes}), communicating over edges (\emph{Topics}). ROS requires a master node to connect to all other nodes, preventing its use in fully decentralized systems. ROS2 is a redesign of ROS that solves the master node issue, enabling completely decentralized systems \cite{ros2_design}. Many popular frameworks have not migrated from ROS to ROS2, preventing their use in fully decentralized multirobot systems. Our software infrastructure leverages ROS2 to create fully independent agents.

\section{Approach}
Our framework can be separated into software and networking infrastructure. In this section, we first explain our software framework. Our framework is capable of running policies in a fully decentralized asynchronous \gls{adhoc} mode, but for the purpose of an experimental ablation analysis, we identify a range of sub-categories with different degrees of decentralization.

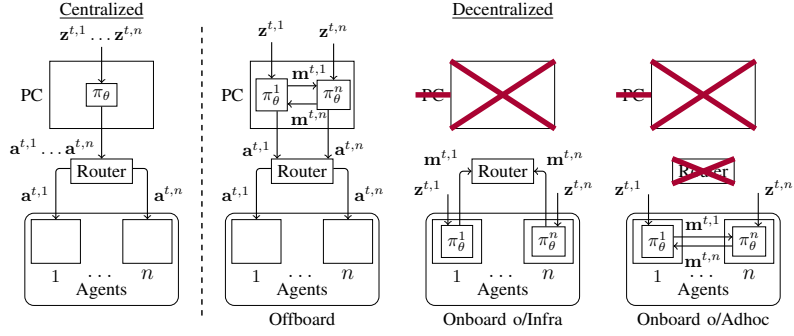
\begin{figure*}[tb]
    \centering
    \resizebox{0.60\textwidth}{!}{\input{tikz_figs/cent}}
    \caption{The framework configurations used in our experiments. The ROS2 infrastructure is either centralized or decentralized, with varying degrees of decentralization depending on the network setup. We refer to these four configurations as Centralized, Offboard, Onboard over Infrastructure, and Onboard over Adhoc. Observations $\bbz^{t,1} \dots \bbz^{t,n}$ feed into centralized policy $\bbpi_\theta$ or local policies $\pi_\theta^1 \dots \pi_\theta^n$ to produce actions $\bba^{t,1} \dots \bba^{t,n}$ for agents $1 \dots n$. Local policies consist of a \Gls{gnn} and pass messages $\bbm^{t,1} \dots \bbm^{t,n}$ to communicate. In the centralized case, a single policy produces actions for all agents at once in a synchronized manner. For \gls{offboard}, local policies run asynchronously, exchanging messages over \texttt{localhost}. The PC is removed for \gls{onboard_infra}, moving inference onto the robot computers. \gls{onboard_adhoc} is fully decentralized -- the agents forgo the router and communicate directly using \gls{adhoc} networking.}
    \label{fig:four_modes}
\end{figure*}

Specifically, we introduce the four modes: \gls{centralized} (fully centralized evaluation), \gls{offboard} (asynchronous evaluation on a central computer), \gls{onboard_infra} (decentralization using existing centralized networking infrastructure) and \gls{onboard_adhoc} (full decentralization using \gls{adhoc} communication networks), as visualized in \autoref{fig:four_modes}. We describe the networking considerations that allow ROS2 to be used for decentralized \gls{adhoc} communication between agents.

\subsection{ROS2 Infrastructure}

\begin{figure}[tb]
    \centering
    \resizebox{0.75\linewidth}{!}{\input{tikz_figs/ros2}}
    \caption{Our ROS2 architecture is composed of the \emph{world} and \emph{agents}. There is one agent $i=1$ in the centralized case, and multiple $i \in \{1 \dots n\}$ in the decentralized case. The agents receive sensor information $\hbz_t$ from either the motion capture system, GPS, or simulator. The aggregator combines sensor information with messages $\bbm^{t, i}$ to produce observation $\bbz^{t, i}$ and neighborhood messages $\bbm^{t, j}; j \in \ccalN^{t,i}$, for the policy $\pi_\theta^i$ to generate action $\bba^{t, i}$. The control node converts the action into velocity commands $\bbv^{t, i}$. In \gls{sim} mode, control drives the simulator instead of the robot wheel motors. The state server orchestrates termination, resets, and operational mode syncs during sequential episodes. This system allows us to run agents in \gls{sim} and the \gls{real_world} concurrently, over multiple episodes, and without any human intervention.}
    \label{fig:ros_diagram}
\end{figure}
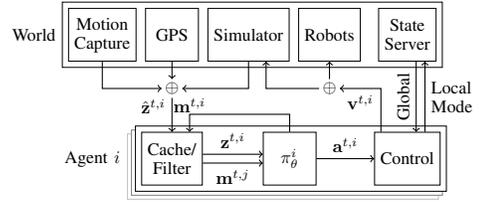

Our multi-agent ROS2 infrastructure (see \autoref{fig:ros_diagram}) allows us to run both simulated and \gls{real_world} agents concurrently, over multiple episodes, in centralized or decentralized mode, and without human intervention (facilitated through automated resets). An episode is one instance of one experimental trial and a reset is a scenario-specific resetting operation, e.g., requiring robots to move to initial positions. These two actions are repeated for a set number of iterations and different initial states. Our infrastructure follows the \gls{rl} paradigm of delineating the \emph{agent} from the \emph{world}.

\subsubsection{Agent}
The agent receives raw sensor data and emits motor commands. The agent is composed of the cache/filter, policy, and control nodes. The cache/filter node uses sensor information $\hbz^{t, i}$ to determine neighboring agents $j\in \ccalN^{t, i}$ within the specified communication radius. It caches neighborhood messages $\bbm^{t, j}$ and sensor information $\bbz^{t, i}$ over $\Delta t$ for the policy. The policy node wraps a trained policy $\pi_\theta^i$. It receives the observation $\bbz^{t, i}$ and messages $\bbm^{t, j}$ and emits a message $\bbm^{t, i}$ and action $\bba^{t, i}$. The action feeds into the control node, which emits motor commands $\bbv^{t, i}$.

\subsubsection{World}
The world is everything external to the agent. The world can be either \emph{real}, \emph{simulated}, or a mix of both. In the \gls{real_world}, an external system like GPS or motion capture produces state estimates for the agents. In the simulated world, a rigid-body dynamics simulator receives agent control commands and moves the agents in \gls{sim} accordingly. All sim-to-real abstraction is contained within the world, so the agents are unaware if they are operating in the \gls{real_world} or the dynamics simulator.

The state server is a state machine that coordinates asynchronous episode execution and resets between independent agents. It enables back-to-back episodes and large-scale experimental data collection. It records agent heartbeats, then broadcasts a global operating mode and initial conditions. Agents use the global operating mode to determine if they should reset or execute the policy.

\subsection{Communications Networks}
Our evaluations consider four different configurations, as summarized in \autoref{fig:four_modes}, which take the form of variable execution locations (i.e., offboard vs onboard) for policies, and the networks used for messaging between agents and policy execution points. \gls{centralized} and \gls{offboard} run policies on an external computer, with the remaining two \gls{onboard} configurations running them on robots' computers. These varied modes allowed us to separate sources of error and performance drop during evaluation.

Our framework uses two wireless communications methods over the various configurations. Both use 802.11, with the first being an \emph{Infrastructure mode} network, and the second being \emph{\gls{adhoc} mode}. We selected 802.11 in preference to other \gls{adhoc} capable wireless standards, such as IEEE 802.15 due to achievable data rates and compatibility with IP-based networking.

\subsubsection{Infrastructure Mode}
\label{sec:infra_mode}
This mode is characterized by a central access point being responsible for managing the network's functions. For \gls{centralized} and \gls{offboard} configurations only agent actions are sent, which is easily handled by the network. For the \gls{onboard_infra} configuration, agents forward messages to one another using this message, with observations from the agent location system sharing the network. Finally, in the \gls{onboard_adhoc} configuration, it handles only the delivery of agent location observations. The implications of each of these modes are discussed further in \autoref{sec:results_network}.

\subsubsection{\gls{adhoc} mode}
We use this network mode only in the \gls{onboard_adhoc} mode, where it handles messages between agents. Physically, this network is supported by distinct wireless transceivers carried with each agent, allowing fully decentralized operation. This network takes the form of an 802.11n IBSS, which means that no agent has any special priority access to the wireless medium. Note that this is \emph{not} a mesh network, as there is no facility for multi-hop communications.

\subsubsection{ROS2 Middleware}
Communications with agents exclusively use ROS2 provided middleware for message passing, specifically the eProsima Fast-DDS implementation of RTPS. Due to the fact that we use dynamic agent discovery rather than setting explicit communications routes, an agent-based firewall is deployed to block RTPS messaging traffic from using the incorrect network interface.

\section{Network Infrastructure}
We investigate the effects of networking, ROS2, and Fast DDS settings on performance.
During evaluations, our primary metric is the probability of delays between packet transmission and reception for message transmission rates between 10 and 500 per second; where the optimal case is all messages delivered with no delay.\footnote{Approximated in the \gls{centralized} and \gls{offboard} configurations}

We carry out all network-specific experiments on a platform of five Raspberry Pis spaced $2$m to $10$m away from each other, in the same lab environment described in \autoref{sec:passage_scenario}. 

\paragraph{Multicast vs Unicast}
We use the eProsima DDS RTPS implementation \cite{GithubRepo_FastDDS}, which defaults to using unicast (one-to-one) communications between publishers and subscribers. This allows reliable transport protocols; however when multiple subscribers are active, the publisher will send duplicate messages as many times as there are subscribers, leading to exponential increases in messaging rates with increasing agent counts. The alternative is to use \gls{multicast}, where each publisher sends only one wireless broadcast for each message. The drawback is that neither reliable transport protocols or the 802.11 hardware based re-transmission mechanism can be utilised, reducing the odds of a given message being delivered.

\paragraph{802.11 hardware retries}
When using 802.11, as the number of competing nodes goes up, the probability of any given packet surviving transmission goes down. This is because if two or more nodes transmit at the same time, both packets are lost, and there is no coordination mechanism. For unicast messages, the lack of an acknowledgement from the receiver will cause re-transmission attempts up to a limit. This limit defaults to 7, and we evaluate the performance of 1, 3, 5 and 7 in our testing. We focus upon lower settings than default because these reduce contention.

\paragraph{Wireless adapter selection and channel bandwidth}
The data rate of the network is dependent upon the distance between participants, transmit power, receiver sensitivity, 802.11 version and channel bandwidth. The final configuration used the Netgear A6210 adapter, based upon the MediaTek MT7612U chipset, and a 40MHz channel width. The adapter was selected because it runs a recent 802.11 version (802.11ac) and had Linux driver compatibility with IBSS mode. This adapter has a maximum transmit power of 18dbm; this is sufficient for ranges as high as 20m between robots even in the presence of interference from neighbouring 802.11 traffic, though communications over 200m were possible in quiet environments.

\paragraph{Reliability}
When RTPS-based reliability is enabled through the ROS2 configuration, subscribers notify publishers when messages are not received as expected through different mechanisms. One of these is the use of positive acknowledgements by subscribers, which allows publishers to re-transmit when messages are lost, but causes subscribers to generate additional packets.

\subsection{Results}
\label{sec:results_network}
\begin{figure}[t]
    \centering
    \includegraphics[width=1\columnwidth]{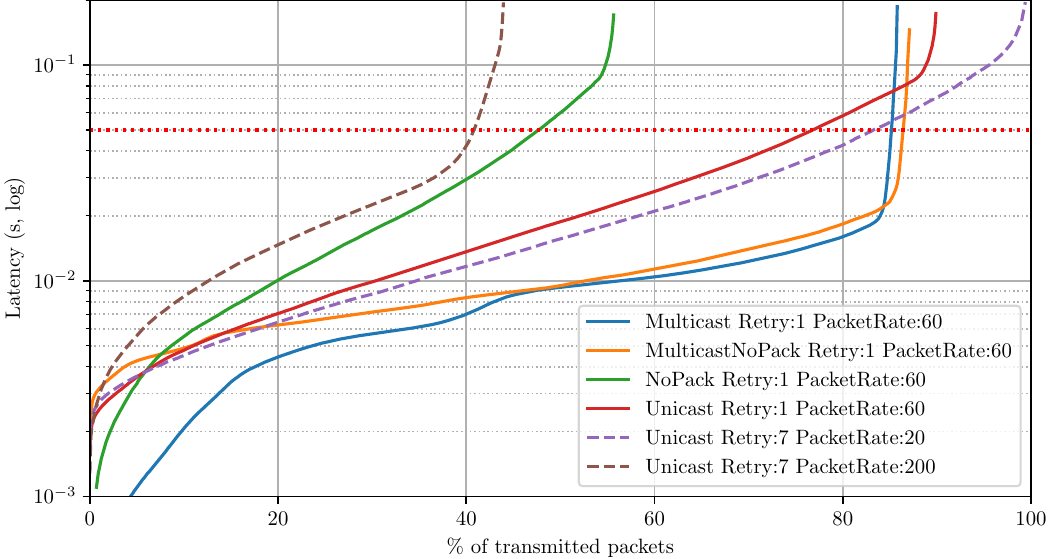}
    \caption{CDF of network evaluation results. y-axis is the delay between one ROS node sending a mesasage via RTPS and the destination receiving it. Four results show Unicast and Multicast performance with and without (NoPack) positive acknowledgements, using only one 802.11 retry at 60 messages per second (approximately the load during episode execution). For context, an additional two results are included which show Unicast performance using seven 802.11 retries at 20 and 200 messages per second, which are default settings. Results are from an 802.11 \gls{adhoc} setup using 5 agents sending messages to every other agent, such that the total rate of messages sent is as specified. The dotted line at 50ms indicates our approximate acceptable limit for latency, above which there is a risk that messages will arrive later than the intended model execution round.}
    \label{fig:comms_result}
\end{figure}

\paragraph{Multicast and 802.11 Retries}
\autoref{fig:comms_result} displays unicast-only performance with dashed lines denoting default 802.11 and RTPS settings at 200 messages per second. With default settings, only 44\% of all messages are delivered, along with consistently higher delays. Even at messaging rates as low as 20 per second, delays remain highly variable and can exceed the interval between policy executions. We found the reduced latency of multicast operation generally performs better than unicast at similar rates despite lower packet delivery rates.
Reducing 802.11 hardware retries to one reduces latency using RTPS unicast defaults (\autoref{fig:comms_result}).

Overall, using multicast, an 802.11 hardware retry setting of one and using RTPS's reliability mechanism results in the most favourable performance, delivering approximately 84\% of messages  within $20$ ms in the \gls{onboard_adhoc} setup, with the remainder being lost.
\autoref{fig:comms_result} highlights the relative latency stability of the chosen scheme where packets are either delivered at low latency, or fail to be delivered at all.

\paragraph{Reliability}
We found disabling positive acknowledgements reduces messaging delays due to the reduction in network contention.

\paragraph{Applicability}
Any adapter conforming to the same 802.11 revision with similar transmit powers and antennas should perform similarly to the presented results. Achievable inter-robot range is limited by wireless interference.
If an increased number of robots are contending for airspace, the total number of packets per second achievable will reduce.

\section{Case Study: Navigation Through Passage}
\label{sec:passage_scenario}
We showcase the capabilities of our framework in a case-study requiring tight coordination between multiple mobile robots. We consider a team of $n=5$ agents that start in a cross-shaped formation and need to move through a narrow passage to reconfigure on the other side of the wall, as seen in \autoref{fig:hero_figure}. The robots are required to reach their goal positions through collision-free trajectories. 
Each robot only has knowledge of its own position and goal (i.e., does \textit{not} directly observe the other robots), and is trained to leverage a \Gls{gnn}-based communication strategy to share this local information with neighbors to find the fastest collision-free trajectory to its respective goal. An image of this setup can be seen in \autoref{fig:hero_figure} and a video demonstration is available online\footref{footnote:video}. We briefly explain the training and provide the code with implementation details online.\footnote{\label{footnote:rl_repository}\rurl{github.com/proroklab/rl_multi_agent_passage}}

\paragraph{Environment}
At each discrete time step $t$, each agent $i$ has a position $\bbp^{t, i}$, a desired velocity $\bbv_d^{i, t}$, a measured velocity $\bbv_m^{i, t}$, and a desired acceleration $\bba_d^{i, t}$.
We approximate each agent to be circular and implement a simple holonomic motion model that integrates acceleration-constrained velocities into positions. Collisions between agents and the wall result in an immediate stop of the agent. Note that the desired velocity is dictated by the control policy and the measured velocity is current true velocity of the agent. Each agent is assigned a goal position $\bbp^{i}_{g}$. An episode ends if all agents have reached their goal or after the episode times out.

\paragraph{Reward}
We train agents using \gls{rl}. The objective of each agent is to reach its goal position $\bbp_{g}^i$ as quickly as possible while avoiding collisions. We use a shaped reward that guides individual agents to their respective goal positions as quickly as possible while penalizing collisions.

\paragraph{Observation and Action}
The observation $\bbz^{i, t}$ consists of locally available information, specifically the absolute position $\bbp^{i, t}$, the relative goal position $\bbp_g^i - \bbp^{i, t}$, as well as a predicted position $\bbp^{t,i} + \bbv^{i, t}$. 
The desired velocity is the policy's action output $\bbv^{i, t}_d = \bba^{i, t}$. We constrain acceleration and velocity to $a_\mathrm{max}=1 \mathrm{m/s}^2$ and $v_\mathrm{max} = 1.5 \mathrm{m/s}$.

\paragraph{Model}
As the model for the policy $\pi_\theta^i$ we use the \Gls{gnn} introduced in \autoref{sec:gnn}. The number of layers is constrained by our communication framework. Since more layers result in multiple rounds of communication exchanges at the same time step, we set $k=1$. Each message is an encoding of the observation so that $\bbm^{t,i} = \bbh_0^{t,i} = \theta_\mathrm{ENC}(\bbz^{t,i})$. We define our message function and vertex update function as $M_1^\theta(\bbh_0^{t,i}, \bbh_0^{t,j}, \cdot) = \theta_\mathrm{GNN}(\bbh_0^{t,i} - \bbh_0^{t,j})$ and $U_1^\theta(\cdot, x) = \theta_\mathrm{ACT}(x)$. Furthermore, we include self-loops and thus consider agent $i$ as part of its own neighborhood so that $\ccalN^{t, i} = \ccalN^{t, i} \cup \{i\}$. The output of the \Gls{gnn} is the desired velocity $\bbv_d^{t, i} = \bba^{t, i} = \bbh_1^{t,i}$. $\theta_\mathrm{ENC}$, $\theta_\mathrm{GNN}$ and $\theta_\mathrm{ACT}$ are learnable \Glspl{mlp}. We use the same approach as described in \cite{blumenkamp_2020} to train our model using PPO with local rewards for each agent. 

\paragraph{Experimental Setup}
In total, we run a series of six different real-world experiments for the four modes (\autoref{fig:four_modes}) to demonstrate the capabilities and performance of our framework and two additional experiments to demonstrate the robustness of our policy against changes to the communication radius in the \gls{real_world}. In addition to using a set communication radius of $R_\mathrm{COM}=2$m, we \emph{(i)} run the policy in a fully connected communication topology, and \emph{(ii)} run the policy in a noisy communication topology by modeling the communication range as a Gaussian with a mean of $R_\mathrm{COM}=2$ m and a standard deviation of $0.5$ m (the policy is trained with $R_\mathrm{COM}=2$ m).

To collect a statistically significant amount of data, we generate $E=16$ episodes, each with a different set of random start and goal positions, and repeat each episode for each experiment $K=12$ times, resulting in $K \cdot E$ episodes in the training environment (\gls{sim}) and on real robots. We use customized DJI RoboMaster robots equipped with Raspberry Pi's that locally run policies. The robots are provided with state information as explained in \autoref{sec:infra_mode}.

\subsection{Results}

\begin{figure*}
\setlength{\tabcolsep}{2pt} 
\newcommand{\figwidth}{2.6cm}
\tiny
\begin{tabular}{ccccc|cc}
& \hspace{0.4cm} \begin{tabular}[c]{@{}l@{}}\gls{centralized}\end{tabular}
& \hspace{0.0cm} \begin{tabular}[c]{@{}c@{}}\gls{offboard}\end{tabular}
& \hspace{0.0cm} \begin{tabular}[c]{@{}c@{}}\gls{onboard_infra}\end{tabular}
& \hspace{0.0cm} \begin{tabular}[c]{@{}c@{}}\gls{onboard_adhoc}\end{tabular}
& \hspace{0.0cm} \begin{tabular}[c]{@{}c@{}}\gls{onboard_adhoc}\\ $R_\mathrm{COM} = \infty$\end{tabular}
& \hspace{0.0cm} \begin{tabular}[c]{@{}c@{}}\gls{onboard_adhoc}\\ $R_\mathrm{COM} = 2 \pm 0.5$ m \end{tabular} \\
\raisebox{0.7cm}[0pt][0pt]{\rotatebox[origin=l]{90}{Makespan}} &
\includegraphics[width=3.25cm]{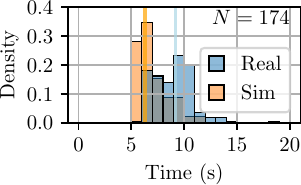} &
\includegraphics[width=\figwidth]{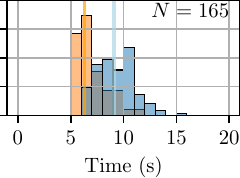} &
\includegraphics[width=\figwidth]{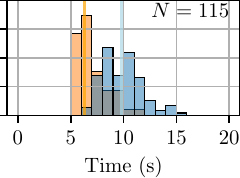} &
\includegraphics[width=\figwidth]{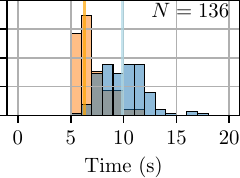} &
\includegraphics[width=\figwidth]{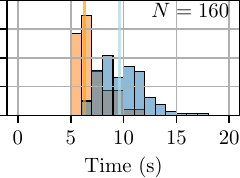} &
\includegraphics[width=\figwidth]{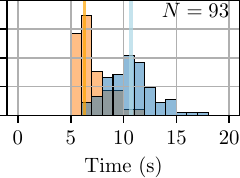} \\
\raisebox{0.95cm}[0pt][0pt]{\rotatebox[origin=l]{90}{$\bbp_x/\bbp_y$}} &
\includegraphics[width=3.2cm]{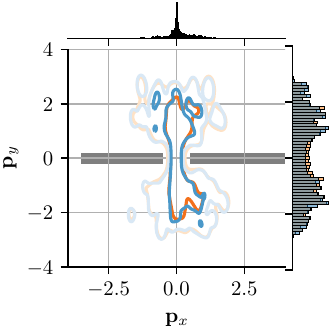} &
\includegraphics[width=\figwidth]{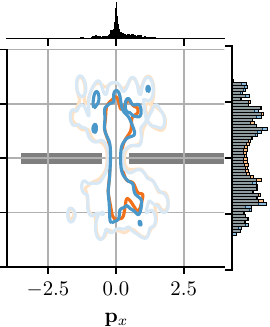} &
\includegraphics[width=\figwidth]{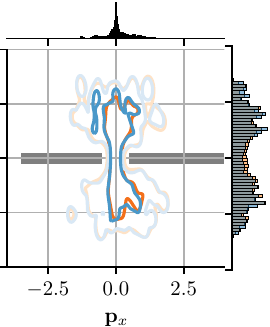} &
\includegraphics[width=\figwidth]{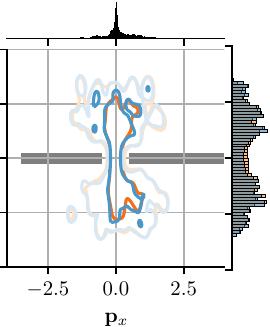} &
\includegraphics[width=\figwidth]{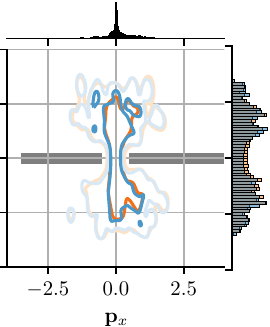} &
\includegraphics[width=\figwidth]{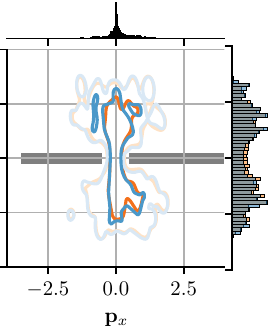} \\
\raisebox{0.7cm}[0pt][0pt]{\rotatebox[origin=l]{90}{$d_\mathrm{min}/d_\mathrm{origin}$}} &
\includegraphics[width=3.0cm]{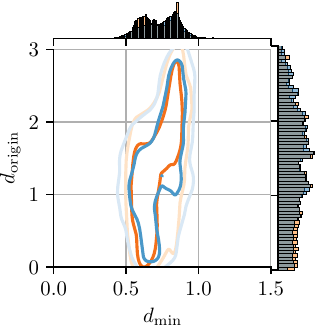} &
\includegraphics[width=\figwidth]{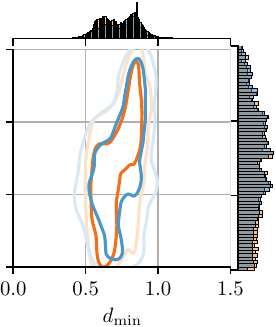} &
\includegraphics[width=\figwidth]{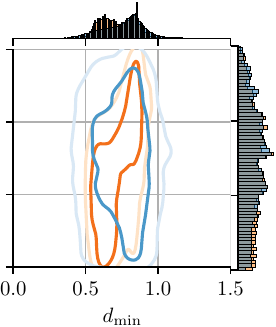} &
\includegraphics[width=\figwidth]{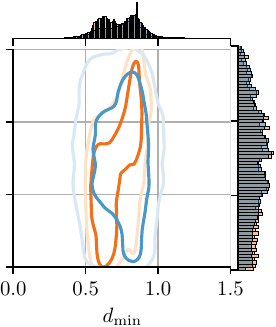} &
\includegraphics[width=\figwidth]{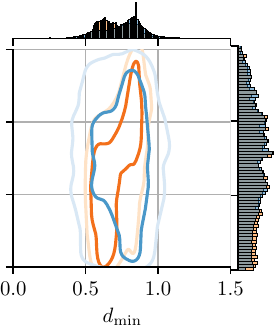} &
\includegraphics[width=\figwidth]{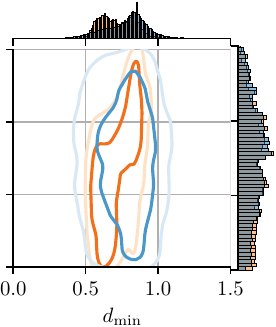} \\
\end{tabular}

\caption{We visualize a variety of makespan and position distributions over the six experiments we conducted. The columns show the data of the centralized \gls{sim} baseline in orange and the data of the corresponding \gls{real_world} experiment as labeled in the column headers in blue. For each experiment, we run a total of $192$ episodes with $16$ different start and goal positions. The last column compares the \gls{onboard_adhoc} experiment with a simulation evaluated with communication delays. The first row shows the probability densities of makespans of successful episodes (episodes that did not result in a collision with the wall and for which all robots reached their goal, indicated with $N$). The median makespan is indicated with a dashed line. The second row shows the distribution of positions, indicating the position of the wall and the passage. The third row shows the distribution of minimum distances between robots at each time step $d_\mathrm{min}$ and distance from the origin or passage $d_\mathrm{origin}$.}
\label{fig:passage_results}
\end{figure*}

\begin{table*}[bt]
\centering
\caption{Overview of performance metrics for all case study experiments.}
\label{tab:passage_results}
\scalebox{0.9}{
\begin{tabular}{cc|cccc|cc}
\toprule
  & \begin{tabular}[c]{@{}l@{}}\gls{sim}\end{tabular}
  & \begin{tabular}[c]{@{}l@{}}\gls{centralized}\end{tabular}
  & \begin{tabular}[c]{@{}c@{}}\gls{offboard}\end{tabular}
  & \begin{tabular}[c]{@{}c@{}}\gls{onboard_infra}\end{tabular}
  & \begin{tabular}[c]{@{}c@{}}\gls{onboard_adhoc}\end{tabular}
  & \begin{tabular}[c]{@{}c@{}}\gls{onboard_adhoc}\\ $R_\mathrm{COM} = \infty$\end{tabular}
  & \begin{tabular}[c]{@{}c@{}}\gls{onboard_adhoc}\\ Noise\end{tabular} \\ \midrule
\input{Figs/results/table}
\bottomrule
\end{tabular}
}
\vspace{-2em}
\end{table*}

We use two metrics to evaluate the performance of our model in \gls{sim} and \gls{real_world}. The \textit{success rate} is the fraction of collision-free episodes for which all robots reached their goal. The \textit{makespan} is the time it takes for the last agent to reach its goal. For both metrics, episodes with wall or inter-agent collisions are excluded. Inter-agent collisions are defined as two agents approaching each other closer than $0.32$ m. We compare to a simulation baseline, for which the policy is evaluated during training conditions. We show distributions of makespans and positions in \autoref{fig:passage_results} and show quantitative results in \autoref{tab:passage_results}.

The \gls{centralized} case reflects the performance gap caused by dynamic constraints that are not considered in \gls{sim}. Since the \Gls{gnn} is evaluated synchronously, communication is not affected by \gls{real_world} effects. The makespan is about $1.5$ times worse and the success rate $5.7$ \gls{pp} worse than in \gls{sim}.

The \gls{offboard} mode evaluates the \Gls{gnn} asynchronously across different processes on the same physical computer. Compared to the \gls{centralized} mode, it features \textit{asynchronous} evaluation but little to no inter-process communication delays, resulting in slightly worse performance of $4.2$ \gls{pp} and worse median makespan of $0.2 s$ wrt the \gls{centralized} mode.

The \gls{onboard_infra} mode moves the decentralized \Gls{gnn} from a central computer to the on-board computers of each individual robot and therefore adds communication delays caused by wireless routing and contention. We notice a decrease in performance of $26.5$ \gls{pp} in terms of success, and a deterioration of $1.0$ s of median makespan w.r.t. the \gls{offboard} mode.
\gls{onboard_adhoc} mode improves the performance by $10.9$ \gls{pp}, with a similar median makespan. This can be attributed to less contention.

Setting $R_\mathrm{COM}=\infty$ results in an identical median makespan and a slight decrease in performance of $3.6$ \gls{pp}. This decrease is expected due to out-of-distribution neighborhoods that never occur during training (while the agents are typically fully connected in the start and the beginning of each episode, they are not when moving through the passage). When adding noise to the communication range, the success rate drops by another $4.2$ \gls{pp} (or $7.8$ \gls{pp} wrt the \gls{onboard_infra} mode) and $0.9$ s median makespan.

The second and third row in \autoref{fig:passage_results} visualize distributions over positions. The second row shows that the distribution of absolute positions over all experiments are consistent, even when comparing to the centralized \gls{sim}. In the third row, we compare the distribution of distance to the origin (or the passage) $d_\mathrm{origin}$ over minimum distance between agents $d_\mathrm{min}$. While the \gls{sim} and real distributions are overlapping in the \gls{centralized} and \gls{offboard} mode, there is a noticeable discrepancy in all Onboard modes, especially for small $d_\mathrm{min}$, for which $d_\mathrm{origin}$ is shifted towards higher values, indicating that the robots are further away from each other when close to the passage, which can be attributed towards slower reaction times caused by communication delays.

We run an additional simulation that evaluates the \gls{gnn} in a decentralized mode with communication delays. We observed that for higher delays, the success rates dropped significantly, while the makespan decreased much less notably. The distribution of $d_\mathrm{min}$ over $d_\mathrm{origin}$ shifted slightly towards the real-world distribution. This indicates that the shift in makespan we observe is mostly due to robot dynamics, and real-world communication latency causes the agents to be less responsive and therefore to collide.

\section{Discussion and Further Work}
This work is the first to demonstrate the \gls{real_world} deployment of a \Gls{gnn}-based policy to a fully decentralized \gls{real_world} multi-robot system using ROS2 and an \gls{adhoc} communication network. We performed a suite of experiments that discuss the selection of suitable networking settings, and subsequently presented results on a real-world scenario requiring tight coordination amongst robots. Our results showed that our framework allows for the successful deployment of our control policy in an \gls{adhoc} configuration, albeit with a performance that is
$22$ \gls{pp} worse in terms of success rate and $9$ \gls{pp} worse in terms of median makespan wrt the centralized mode.

Even though the deployment of our scenario was successful, we reported a degradation of performance when moving from simulation to the real world, which can be attributed to \gls{real_world} effects such as communication delays.
In the future, we plan to use our software framework to validate novel mechanisms that are robust to communications-specific domain shifts and thus aid in closing the sim-to-real gap for \Glspl{gnn}.
We believe that the presented framework will facilitate the deployment of robot systems into more complex environments and the unstructured outdoors, potentially leveraging more complex networking architectures such as mesh networks and on-board sensing.

\footnotesize{
\section*{Acknowledgment}
J. Blumenkamp acknowledges the support of the `Studienstiftung des deutschen Volkes' and an EPSRC tuition fee grant. 
We gratefully acknowledge the support of ARL grant DCIST CRA W911NF-17-2-0181, the EPSRC (grant EP/S015493/1), and ERC Project 949940 (gAIa). The support of Arm is gratefully acknowledged. 
}

\addtolength{\textheight}{-2.9cm}
\clearpage
\bibliographystyle{IEEEtran}
\bibliography{IEEEabrv, bibliography}

\end{document}

%% file: definitions.tex
\renewcommand{\equationautorefname}{\fixit{use \textbackslash eqref for equations instead of \textbackslash autoref!}}

\newacronym{gnn}{GNN}{Graph Neural Network}
\newacronym{mlp}{MLP}{Multilayer Perceptron}
\newacronym{rl}{RL}{Reinforcement Learning}
\newacronym{il}{IL}{Imitation Learning}
\newacronym{ros}{ROS}{Robot Operating System}
\newacronym{pp}{pp}{percentage points}

\newglossaryentry{centralized}{
    name=Centralized,
    description={},
}
\newglossaryentry{offboard}{
    name=Offboard,
    description={},
}
\newglossaryentry{onboard}{
    name=Onboard,
    description={},
}
\newglossaryentry{onboard_infra}{
    name=Onboard o/Infra,
    description={},
}
\newglossaryentry{onboard_adhoc}{
    name=Onboard o/Adhoc,
    description={},
}
\newglossaryentry{multicast}{
    name=Multicast,
    description={},
}
\newglossaryentry{real_world}{
    name=real-world,
    description={},
}
\newglossaryentry{sim}{
    name=simulation,
    description={},
}
\newglossaryentry{adhoc}{
    name=Adhoc,
    description={},
}

\defglsentryfmt{%
  \ifglsused{\glslabel}{%
    \glsgenentryfmt%
  }{%
    \emph{\glsgenentryfmt}%
  }%
}


%% file: tikz_figs/hero.tex
\begin{tikzpicture}[auto,>=latex']
    \def\centerarc[#1](#2)(#3:#4:#5)
        { \draw[#1] ($(#2)+({#5*cos(#3)},{#5*sin(#3)})$) arc (#3:#4:#5); }
    \tikzset{
        enc/.style = {trapezium, trapezium angle=-75, inner sep=0.1em, draw},
        wheel/.style = {draw=black, fill=lightgray},
        wall/.style = {draw=black, pattern=north west lines},
        pics/enc/.style args={#1}{code={
            \node [enc] (_e) {$\theta_\mathrm{ENC}$};
            \node [coordinate, above=0.4em of _e] (_in) {};
            \node [coordinate, below=1em of _e] (_out) {};
            \node [inner sep=0, above=0.1em of _in] (_in_lbl) {$\bbz^{t,#1}$};

            \draw[->](_in) -- (_e);
        }},
        pics/robot/.style args={#1}{code={
            \node[circle, draw, minimum size=#1*1em, inner sep=0] (_body) {};
            \node[coordinate] (_origin) {};

            \def\wheeldist{0.35em}
            \def\wheelbase{0.2em}
            \def\wheelheight{0.125em}
            \def\wheelwidth{0.25em}
            \draw[wheel] ++(#1*\wheeldist,#1*\wheelbase) rectangle ++(#1*\wheelheight,#1*\wheelwidth);
            \draw[wheel] ++(#1*-\wheeldist,#1*\wheelbase) rectangle ++(-#1*\wheelheight,#1*\wheelwidth);
            \draw[wheel] ++(#1*\wheeldist,#1*-\wheelbase) rectangle ++(#1*\wheelheight,-#1*\wheelwidth);
            \draw[wheel] ++(-#1*\wheeldist,-#1*\wheelbase) rectangle ++(-#1*\wheelheight,-#1*\wheelwidth);
            
        }},
        pics/agent/.style args={#1}{code={
            \pic[yshift=-0.3em] (_enc) {enc=#1};
            \pic (_robot) {robot={4}};
        }},
        relative to node/.style={
            shift={(#1.center)},
            x={(#1.east)},
            y={(#1.north)},
        },
        comm/.style = {decorate,decoration={snake, amplitude=0.4mm, segment length=3mm, post length=1mm}},
    }
    \node [circle, draw, inner sep=0.1em] (agg) {$\sum$};
    \pic[above=1.6em of agg] (enc_i) {enc={i}};
    \draw[-](enc_i_e.south) -- (enc_i_out.south);
    \node [inner sep=0, above left=0em and 0.1em of enc_i_out] {$\bbm^{t,i}$};
    
    \foreach \angle / \agent / \dist in {198/j/7.3em, 163/k/6.5em, -5/l/7em} {
        \path (agg) ++(\angle:3.5em) node (m_\agent) [circle, draw, inner sep=0em] {$M_1^\theta$};
        \pic[above right=cos(\angle-90)*\dist and sin(\angle+90)*\dist of m_\agent.center, anchor=center, rotate fit=\angle + 270, transform shape] (agent_\agent) {agent={\agent}};
        
        \draw[->, comm](agent_\agent_enc_e.south) -- node[pos=0.5,above,sloped] {$\bbm^{t,\agent}$} (m_\agent);
                
        \draw[->](enc_i_out.south) -- (m_\agent);
        \draw[->](m_\agent) -- (agg);
    }

    \foreach \angle/\agent/\dist in {12/a/15.3em} {
        \pic[above right=cos(\angle-90)*\dist and sin(\angle+90)*\dist of agg.center, anchor=center, rotate fit=\angle + 270, transform shape] (agent_\agent) {agent={\agent}};
        \draw[->, comm](agent_\agent_enc_e.south) -- (agent_\agent_enc_out);
        \node [inner sep=0, left=1.6em of agent_\agent_enc_out, anchor=center] {$\bbm^{t,\agent}$};
    }

    \pic[rotate fit=90, transform shape] (robot_i) at (agg) {robot={10.5}};
    \node[below=4.5em of robot_i_origin, anchor=center] {\tiny Close-Up \reflectbox{\faSearchPlus}};
    \faSearch

    \node[below=0.7em of agg, circle, draw, inner sep=0.1em] (u) {$U_1^\theta$};
    \draw[->](agg) -- (u);
    \node[right=0.7em of u, inner sep=0.1em] (v) {$\bbv_d^{t, i}$};
    \draw[->](u) -- (v);

    \draw[wall] (agg) ++(7em,2em) rectangle ++(2em,3.5em);
    \draw[wall] (agg) ++(7em,-4em) rectangle ++(2em,-1.5em);
    
    \def\commrange{13em}
    \def\commrangeangle{27}
    \centerarc[dashed](agg)(180-\commrangeangle:180+\commrangeangle:\commrange);
    \centerarc[dashed](agg)(\commrangeangle:-\commrangeangle:\commrange);
    
    \node[below right=4.5em and 13em of agg, anchor=center] {$R_\mathrm{COM}$};
    \node[below left=-0.5em and 9em of agg, anchor=center] {$\mathcal{N}^{t, i} = \{j, k, l\}$};

\end{tikzpicture}

%% file: tikz_figs/cent.tex
\begin{tikzpicture}
    \tikzset{
        pre/.style={=stealth',semithick},
        post/.style={->,shorten >=1pt,>=stealth',semithick},
        pics/frame/.style args={#1}{code=
            {
                \node[minimum height=3em, minimum width=5em] (anchor) {};
                \node[draw, fit=(anchor), label={[name=pc_box_label]left:PC}] (pc_box) {};
                \node[draw, below=0.7cm of anchor] (router) {Router};
                \node[minimum height=1.7em, minimum width=2em, below left=3em and 1.5em of router.center] (agent_anchor_left) {};
                \node[draw, fit=(agent_anchor_left), label={[name=l0] below: $1$}] (agent_box_left) {};
                
                \node[minimum height=1.7em, minimum width=2em, below right=3em and 1.5em of router.center] (agent_anchor_right) {};
                \node[draw, fit=(agent_anchor_right), label={[name=l1] below: \large $n$}] (agent_box_right) {};
                \node[below=4.65em of router] (dots) {\large $\dots$};
                \node[fit=(agent_box_left) (agent_box_right) (l0) (l1), rounded corners=5pt] (all_agents) {};
                \node[align=center, below=0.01em of dots, inner sep=0pt, outer sep=0pt] (agent_text) {Agents};
                \node[draw, fit=(agent_box_left) (agent_box_right) (l0) (l1) (agent_text), rounded corners=5pt] (all_agents_box) {};

                \node[below=0.01em of all_agents_box] (fig_index) {#1};

            }
        }
    }
    
    \pic[] (cent_) {frame={}};
    \node[draw] (cent_inf) at (cent_anchor.center) {$\mathbb{\pi}_\theta$};
    \draw[->, semithick] (cent_inf) -- (cent_router) node[midway, below left] {$\bba^{t,1} \dots \bba^{t,n}$};
    \node[above=2em of cent_inf] (cent_pos) {$\bbz^{t,1} \dots \bbz^{t,n}$};
    \draw[->, semithick] (cent_pos) -- (cent_inf);
    
    \draw[->, rounded corners=2pt, semithick] (cent_router.175)-|(cent_agent_box_left) node[near end, left] {$\bba^{t,1}$};
    \draw[->, rounded corners=2pt, semithick] (cent_router.5)-|(cent_agent_box_right) node[near end, right] {$\bba^{t,n}$};
    \node[above=2.3em of cent_anchor] {\underline{Centralized}};

    \pic[right=6em of cent_anchor] (pc_cent_) {frame={Offboard}};

    \node[draw, left=0.8em of pc_cent_anchor.center] (pc_cent_inf) {$\pi_\theta^1$};
    \node[draw, right=0.8em of pc_cent_anchor.center] (pc_cent_inf2) {$\pi_\theta^n$};
    \draw[->, semithick] ([yshift=0.5em]pc_cent_inf.east) -- ([yshift=0.5em]pc_cent_inf2.west) node[near end, above] {$\bbm^{t,1}$};
    \draw[<-, semithick] ([yshift=-0.5em]pc_cent_inf.east) -- ([yshift=-0.5em]pc_cent_inf2.west) node[near end, below] {$\bbm^{t,n}$};

    \draw[->, semithick] ([xshift=0.3em]pc_cent_inf.south) -- ([xshift=0.3em]pc_cent_router.north -| pc_cent_inf) node[midway, below left] {$\bba^{t,1}$};
    \draw[->, semithick] ([xshift=-0.3em]pc_cent_inf2.south) -- ([xshift=-0.3em]pc_cent_router.north -| pc_cent_inf2) node[midway, below right] {$\bba^{t,n}$};
    
    \node[above=2em of pc_cent_inf] (pc_cent_pos) {$\bbz^{t,1}$};
    \draw[->, semithick] (pc_cent_pos) -- (pc_cent_inf);
    
    \node[above=2em of pc_cent_inf2] (pc_cent_pos2) {$\bbz^{t,n}$};
    \draw[->, semithick] (pc_cent_pos2) -- (pc_cent_inf2);
    
    \draw[->, rounded corners=2pt, semithick] (pc_cent_router.175)-|(pc_cent_agent_box_left) node[near end, left] {$\bba^{t,1}$};
    \draw[->, rounded corners=2pt, semithick] (pc_cent_router.5)-|(pc_cent_agent_box_right) node[near end, right] {$\bba^{t,n}$};
    
    \draw[-, thick, dashed] ($(cent_fig_index)!0.5!(pc_cent_fig_index)$) -- ([yshift=16em]$(cent_fig_index)!0.5!(pc_cent_fig_index)$);

    \pic[right=6em of pc_cent_anchor] (decent_) {frame={Onboard o/Infra}};
    \node[draw] (decent_inf0) at (decent_agent_box_left) {$\pi_\theta^1$};
    \node[draw] (decent_inf1) at (decent_agent_box_right) {$\pi_\theta^n$};
    \draw[<-, rounded corners=2pt, semithick] (decent_router.west)-|([xshift=0.15em]decent_inf0.north) node[near start, above left] {$\bbm^{t,1}$};
    \draw[<-, rounded corners=2pt, semithick] (decent_router.east)-|([xshift=-0.15em]decent_inf1.north) node[near start, above right] {$\bbm^{t,n}$};
    
    \draw[->, semithick] ([xshift=-0.5em,yshift=3.5em]decent_inf0.south) -- ([xshift=-0.5em]decent_inf0.north) node[near start, above left]{$\bbz^{t,1}$};
    
    \draw[->, semithick] ([xshift=0.5em,yshift=3.5em]decent_inf1.south) -- ([xshift=0.5em]decent_inf1.north) node[near start, above right]{$\bbz^{t,n}$};

    \draw[-, line width=3pt, red] (decent_pc_box.south west) -- (decent_pc_box.north east);
    \draw[-, line width=3pt, red] (decent_pc_box.south east) -- (decent_pc_box.north west);
    \draw[-, line width=3pt, red] (decent_pc_box_label.east) -- (decent_pc_box_label.west);

    \node[above=2.3em of decent_anchor] {\underline{Decentralized}};

    \pic[right=6em of decent_anchor] (adhoc_) {frame={Onboard o/Adhoc}};

    \node[draw] (adhoc_inf0) at (adhoc_agent_box_left) {$\pi_\theta^1$};
    \node[draw] (adhoc_inf1) at (adhoc_agent_box_right) {$\pi_\theta^n$};
    \draw[->, semithick] ([yshift=0.25em]adhoc_inf0.east) -- ([yshift=0.25em]adhoc_inf1.west) node[midway, above] {$\bbm^{t,1}$};
    \draw[<-, semithick] ([yshift=-0.25em]adhoc_inf0.east) -- ([yshift=-0.25em]adhoc_inf1.west) node[midway, below] {$\bbm^{t,n}$};
    
    \draw[->, semithick] ([xshift=-0.5em,yshift=3.5em]adhoc_inf0.south) -- ([xshift=-0.5em]adhoc_inf0.north) node[near start, above left]{$\bbz^{t,1}$};

    \draw[->, semithick] ([xshift=0.5em,yshift=3.5em]adhoc_inf1.south) -- ([xshift=0.5em]adhoc_inf1.north) node[near start, above right]{$\bbz^{t,n}$};

    \draw[-, red, line width=3pt] (adhoc_pc_box.south west) -- (adhoc_pc_box.north east);
    \draw[-, red, line width=3pt] (adhoc_pc_box.south east) -- (adhoc_pc_box.north west);

    \draw[-, red, line width=3pt] (adhoc_router.south west) -- (adhoc_router.north east);
    \draw[-, red, line width=3pt] (adhoc_router.south east) -- (adhoc_router.north west);
    
    \draw[-, line width=3pt, red] (adhoc_pc_box_label.east) -- (adhoc_pc_box_label.west);
\end{tikzpicture}

%% file: tikz_figs/ros2.tex
\begin{tikzpicture}
    \tikzset{    
        cascaded/.style = {%
            general shadow = {%
                shadow scale = 1,
                shadow xshift = -1ex,
                shadow yshift = -1ex,
                draw=lightgray,
                fill = white,
            },
            general shadow = {%
                shadow scale = 1,
                shadow xshift = -.5ex,
                shadow yshift = -.5ex,
                draw=gray,
                fill = white
            },
            draw,
            fill = white
        },
        node distance = 1em and 0.25em,
        box/.style = {%
            draw, 
            minimum width=3em, 
            minimum height=3em,
            align=center
        }
    }
    \node[box] (mocap) {Motion\\Capture};
    \node[box, right=0.5em of mocap] (gps) {GPS};
    \node[box, right=0.5em of gps] (sim) {Simulator};
    \node[box, right=0.5em of sim] (robot) {Robots};
    \node[box, right=1em of robot, align=center] (server) {State\\Server};

    \node[draw, fit=(mocap) (sim) (server) (robot), label={[anchor=east]west:World}] (world) {};

    \node[box, below=4em of gps] (filcher) {Cache/\\Filter};
    \node[box, below=4em of server] (ctrl) {Control};
    \node[box] (policy) at ($(filcher)!0.5!(ctrl)$) {$\mathbb{\pi}^i_\theta$};

    \node[draw, fit=(filcher) (policy) (ctrl), label={[label distance=0.5em]left:Agent $i$}, cascaded] (agent) {};
    
    \node[box, below=4em of gps] (filcher) {Cache/\\Filter};
    \node[box, below=4em of server] (ctrl) {Control};
    \node[box] (policy) at ($(filcher)!0.5!(ctrl)$) {$\mathbb{\pi}^i_\theta$};
    
    \node[outer sep=0.1em, inner sep=0.1em, below=of gps] (z_junc) {$\oplus$};
    \node[outer sep=0.1em, inner sep=0.1em, below=of robot] (ctrl_junc) {$\oplus$};

    \draw[->, semithick] ([yshift=0.25em]filcher.east) -- ([yshift=0.25em]policy.west) node[midway, above] {$\bbz^{t,i}$} ;
    \draw[->, semithick] ([yshift=-0.25em]filcher.east) -- ([yshift=-0.25em]policy.west) node[midway, below] {$\bbm^{t,j}$};

    \draw[->, semithick] (policy) -- (ctrl) node[midway, above] {$\bba^{t,i}$};

    \draw[->, semithick] (mocap.south) |- (z_junc.west);
    \draw[->, semithick] (gps.south) -- (z_junc.north);
    \draw[->, semithick] (sim.south) |- (z_junc.east);
    \draw[->, semithick] (z_junc.south) -- (filcher.north) node[near start, left] {$\hbz^{t,i}$};
    
    \draw[<-, semithick] ([xshift=0.5em]ctrl.north) -- ([xshift=0.5em]server.south) node[midway, above, rotate=90] {Global};
    \draw[->, semithick] ([xshift=1.0em]ctrl.north) -- ([xshift=1.0em]server.south) node[midway, below, text width=1cm,anchor=west] {Local\\Mode};
    
    \draw[->, semithick] (ctrl_junc) -| ([xshift=2em]sim);
    \draw[->, semithick] (ctrl_junc) -- (robot);
    \draw[->, semithick] ([xshift=-1.5em]ctrl.north) |- (ctrl_junc) node[near end, below] {$\bbv^{t,i}$};
    
    \draw[->, semithick] (policy.north) |- ([xshift=1em, yshift=1em]filcher.north) -- ([xshift=1em]filcher.north) node[near start, above] {$\bbm^{t,i}$};
\end{tikzpicture}

%% file: Figs/results/table.tex
Success Rate & $95.8$\% & $90.1$\% & $85.9$\% & $59.4$\% & $70.3$\% & $66.7$\% & $62.5$\% \\
Median Makespan & $6.3$ s & $9.1$ s & $9.1$ s & $9.8$ s & $9.9$ s & $9.6$ s & $10.7$ s \\